\documentclass[twoside]{article}

\usepackage[accepted]{aistats2024}
\usepackage{natbib}
\usepackage{graphicx}
\usepackage{amsmath}
\usepackage[ruled,vlined]{algorithm2e}
\usepackage[utf8]{inputenc} 
\usepackage[T1]{fontenc}    
\usepackage{hyperref}       
\usepackage{url}            
\usepackage{booktabs}       
\usepackage{amsfonts}       
\usepackage{nicefrac}       
\usepackage{microtype}      
\usepackage{xcolor}         
\usepackage{isomath}
\usepackage{breqn}
\usepackage{amsbsy}
\usepackage{bm}

\begin{document}

%
\runningtitle{Composite Survival Analysis}

%
\runningauthor{Composite Survival Analysis}

\twocolumn[

\aistatstitle{Composite Survival Analysis: Learning with Auxiliary Aggregated Baselines and Survival Scores}

\aistatsauthor{Chris Solomou *}

\aistatsaddress{ University of York}]

\begin{abstract}
Survival Analysis (SA) constitutes the default method for time-to-event modeling due to its ability to estimate event probabilities  of sparsely occurring events over time. In this work, we show how to improve the training and inference of SA models by decoupling their full expression into (1) an aggregated baseline hazard, which captures the overall behavior of a given population, and (2) independently distributed survival scores, which model idiosyncratic probabilistic dynamics of its given members, in a fully parametric setting. The proposed inference method is shown to dynamically handle right-censored observation horizons, and to achieve competitive performance when compared to other state-of-the-art methods in a variety of real-world datasets, including computationally inefficient Deep Learning-based SA methods and models that require MCMC for inference. Nevertheless, our method achieves robust results from the outset, while not being subjected to fine-tuning or hyperparameter optimization.
\end{abstract}
\vspace{-.25cm}

\section{Introduction}
\vspace{-.15cm}
Survival Analysis is a statistical method used to predict the time until a specific event of interest occurs within a predefined period. Some of its applications include clinical trials, digital marketing, financial forecasting, engineering and manufacturing. Although these domains differ significantly, the underlying purpose of Survival Analysis remains the same, with the event being adjusted to fit the specific domain. Across all applications of Survival Analysis, the objective is to determine the distribution of survival times, which represents significant events such as patient mortality (clinical trials), conversion of a new customer (digital marketing) or failure of a machine (manufacturing). One of the most prominent challenges in Survival Analysis is that over the predetermined period of a study, participants may leave the study early due to a different event than the one of interest. Nonetheless, for many of the participants, that event can occur at a later time, after the study has concluded. This phenomenon represents 'right-censoring' and is commonly encountered in various applications of Survival Analysis.
\vspace{-.22cm}
\begin{figure}[htb]
\centering
\includegraphics[width=2.45in,height=2.7in]{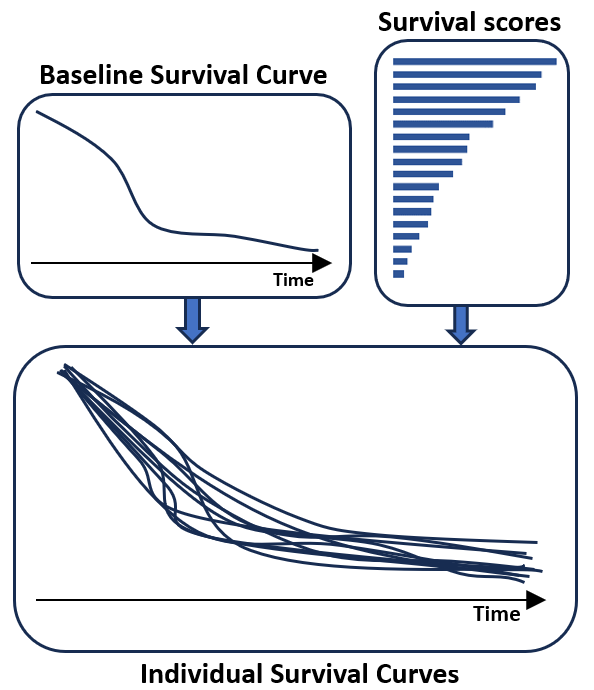}
\vspace{-.1cm}
\end{figure}
\vspace{-.23cm}

\textbf{Related Work} The idiosyncrasies of the event probabilities changing with time, and participants leaving or outrunning the duration of the study, make common classifiers ineffective in Survival Analysis. Common solutions for estimating the survival probabilities over time include methods that take into account specific characteristics of the population (covariates); while others focus specifically on the time-to-event relationship. There is a rich history of non-parametric models applied to survival analysis with the most popular being the Kaplan Meier estimator, developed by  \citet{kaplan1958nonparametric}, which is a non-parametric method for modelling the survival function as the population of the study changes with time. We direct the reader to \citet{fong2022predictive} for more resources of non-parametric methods applied to Survival Analysis. 

\begin{figure*}[h!]
\centering
\includegraphics[width=6.35in,height=1.625in]{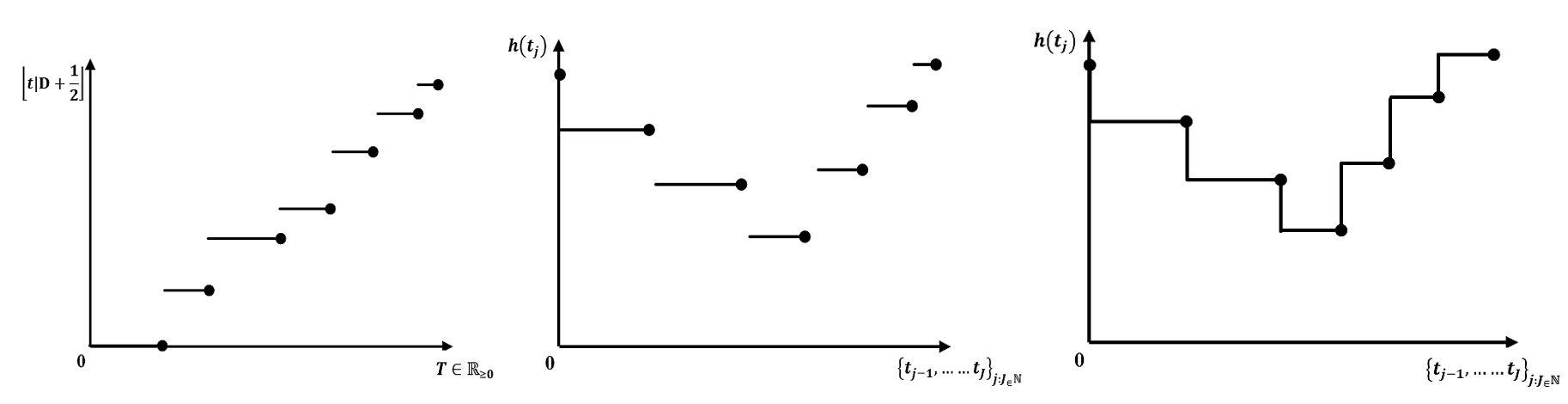}
\vspace{-.3cm}
\caption{\small{\textbf{The discretization of time $(T)$, defined by continuous r.v. Each point in time (from the real line) that an event from the state space $\bm{\{0,1}\}$ has occurred, that observation is rounded to the nearest integer. This results in discretizing $(T)$ and generating the time interval for each step (left). The figure on the right shows a paradigm of the bathtub hazard curve. The approach allows for identifying the time steps for which the hazard rate changes, and subsequently the intervals that the hazard rate remains constant.}}}
\end{figure*}
\vspace{-.2cm}

Traditional parametric methods include modelling the survival times using the Exponential, Gamma, Weibull distribution etc. These methods require a distributional assumption, and it’s difficult to account for the covariates of a study with the few parameters that govern them. Although when their assumptions are met they can handle all types of censored data, and as demonstrated by \citet{efron1977efficiency} and \citet{oakes1977asymptotic} a parametric model yields more efficient estimates than semi-parametric models.

Survival models that utilize covariates include parametric distributions Accelerated Life Models, and Proportional Hazard Models. The \citet{cox1972regression} proportional hazards model, constitutes the most popular model in survival analysis since no assumption is required about the probability distribution of the survival times. It is comprised from a non-parametric part (baseline hazard) and a parametric part that utilizes the population's covariates for modelling the hazard rate over time. 

Machine Learning algorithms have also found success in Survival Analysis. This includes Survival Support Vector Machines (\citet{polsterl2015fast,polsterl2016efficient}), an extension of the standard SVM - \citet{boser1992training}, applied to right-censored time-to-event data. Survival SVM can effectively model non-linear relationships between features and survival outcomes by utilizing the kernel trick. The kernel function transforms the input features into high-dimensional feature spaces and separating hyperplane determines survival (or otherwise). 

Another prevalent method for survival analysis is Random Survival Forest by \citet{ishwaran2008random} an extension of \cite{breiman2001random}. In RSF each tree is grown by randomly selecting a subset of variables at each node and then splitting the node using a survival and censoring criterion. 

More recent work from \citet{muse2021bayesian} proposes a PH approach by utilizing a Generalized Log-Logistic (GLL) model for predicting the baseline hazard, and Bayesian Inference for estimating the regression coefficients. \citet{ranganath2016deep} use a generative approach for jointly modeling the observations and covariates using DEFs \citet{ranganath2015deep} a class of latent variable models that make the distributional assumption that the data can be modelled exponentially. \citet{miscouridou2018deep} expand on this concept and develop flexible survival likelihoods by using deterministic transformations that map the true failure distribution to other distributions, thus allowing to construct invertible mappings. \citet{fernandez2016gaussian} propose a Bayesian model that is centered on a parametric baseline hazard and a non-parametric component for modelling variations. 

The prominence of Deep Learning has also extended to the field of Survival Analysis,  with many DL models achieving state of the art results. The first successful application of DL in Survival Analysis comes from \citet{katzman2018deepsurv} where the parametric part of the Cox Model is replaced by a Neural Network (DeepSurv), although still constrained by the proportionality assumption. \citet{kvamme2019time} overcome this limitation by proposing a loss function that scales well both for the proportional and non-proportional cases. Similar works inspired by Cox regression include work by \citet{yousefi2017predicting} includes SurvivalNet, a framework for fitting proportional Cox models with Neural Networks. We direct the reader to \citet{kvamme2019time} for more references on the applications of DL in Survival Analysis.

New approaches to Survival Analysis, include the application of Reinforcement Learning to time-to-event data. \cite{maystre2022temporally} combined temporal-difference learning, a central concept in reinforcement learning, with Survival Analysis, constructing a 'pseudo-target' that integrates the next-step transition with a prediction about survival.

Analyzing the current and past work, it can be seen that there isn't a standard approach for performing Survival Analysis. Some methodologies focus purely on the time-to-event relationship or the overall discriminative performance; making them unable to produce specific survival probabilities in the form of survival curves for each member of the population. Nonetheless, many approaches are based on computationally demanding and data-intensive DL models or require time-consuming MCMC for inference.

In this work, we aim to address these limitations and present a novel approach to Survival Analysis. Our proposed method exhibits speed, flexibility, and robust performance, even when applied to small datasets. Our contributions can be summarized as follows:
\vspace{.125cm} \\ 
1. We introduce a methodology for decoupling the survival function into two components: the baseline survival and survival scores. This framework can accommodate any baseline survival function and various techniques for deriving survival scores.
\vspace{.125cm} \\
2. We derive a parametric baseline survival function that captures the survival behavior of a population and enables dynamic handling of right-censoring.
\vspace{.125cm} \\
3. We generate individual survival scores and scale them with the baseline survival to create survival curves for each population member.
\vspace{-.22cm}
\section{Background}
\vspace{-.175cm}
Survival Analysis aims to generate survival probabilities as a function of time for a group of participants that comprise a study of interest. We define time $(T)$ to be a continuous r.v., $T \in {\mathbb{R}_{\geq 0}}$ that lasts the duration of the study. In practice we are interested in identifying how the distribution of $(T)$ is affected conditioned on a set of some known covariates, i.e.: $X = \{X_1,X_2, ..., X_n\}$. These covariates can be continuous or discrete and represent the scope of the study in question. 
In turn, the survival probabilities can be defined as a function of: $S(t|X) = P(T > t|X) = 1 - F(t|X)={\int{_{t}^{\infty}}} f(T|X) \, dT$  which stands for the probability that the event of interest has not occurred by some time $t$. However, in a real-world application, the event times are often not observed by all participants, due to right-censoring. Additionally, the joint integral would have to be estimated for $n$ covariates, which often leads to no analytical solutions. In this work, we solve these problems by taking a different approach for inferring the survival function of a population.

We define a member of the study as a combination of the particular values assumed by the covariates, i.e. $x = (X_1 = x_1, X_2 = x_2,...\,, X_n = x_n)$, with $D \in \{0,1\}$ representing the occurrence of the event of interest, and $T^* \in T$ the last recorded time of each member. Thus the observed data from a typical study are given as a set of the form $\{(x_i,T^*_i,D_i)\}_{i=1}^N$. Particularly we aim to utilize this set of data to model a global baseline survival function and individual survival scores for each member of the study. Thereafter, the individual survival curves for each member are generated. This can be achieved by constructing a model that utilizes known covariates and predicts survival probabilities for new participants based on an observed training set ${\{(x_i,T^*_i,D_i)}\}_{i=1}^{N_{train}}$.

Specifically, we are interested in modelling the probability of the occurrence of an event $D \in \{0,1\}$  over a predetermined time interval. Although in classical Survival Analysis $(T)$ is considered a continuous variable, here we discretize this period into time intervals $\{t_{j-1},...\,,t_J\}_{j:J \in \mathbb{N}}$ where \small $J$ \normalsize refers to the end of the study, and $t_j$ to the cardinal value of the set of times up to time \small $J$ \normalsize including 0. Since we want to estimate the hazard rate over a known interval of time, a piecewise constant function can be implemented for generating the set of times where the event of interest has occurred.
\begin{equation*}
\sup \left\{ (t|D) \in T_{\mathbb{R}_{\geq 0}} : \lfloor (t|D) + \frac{1}{2} \rfloor
\leq \lfloor T + \frac{1}{2} \rfloor \right\}
\tag{2.1}
\end{equation*}

\section{The Baseline Hazard Function}
\vspace{-.2cm}
In this section, we introduce our methodology for estimating the baseline hazard and survival functions. We make the following assumptions, and adopt a Bayesian approach to determine the hazard rate at each time interval.

\textbf{Assumption 1:} \textit{Independence - Covariates are independent w.r.t. each other. Formally, for a collection of observations $\{x_i\}_1^n \in \mathbb{R}$ covariates and a given future event occurrence $D \in \{0,1\}$}:
\vspace{-.15cm}
\begin{equation*}
\centering
P(D|X_1 = x_1,…,X_n = x_n) = \prod_{i=1}^n P(D|X_i = x_i)
\end{equation*}
\textbf{Assumption 2:} \textit{Noramlly Distributed - Covariates $\{X_i\}_1^n \in \mathbb{R}^N$ follow a Normal (Gaussian) distribution. In particular, for a single covariate $X$ of size $N$, if $N \rightarrow\ \infty$ , then $X$ can be approximately normally distributed s.t. $X\sim N(\mu,\sigma^2)$.}

We leverage the above assumptions and apply Bayes' principle to each individual covariate. In particular, Assumption 1 allows to optimize each covariate separately, while Assumption 2 enables the aggregation of covariates into a single probability distribution, facilitating sampling from the resulting distribution. 

Specifically, for a set of $n$ covariates and a sequence of times $\{t_{j-1}, t_j, \ldots, t_J\}$, we perform Bayesian inference to estimate the probability of the event of interest (i.e., the hazard rate) at each time step.
\begin{equation*}
\centering 
h(t_j) = \prod_{i = 1}^n P(\mu_{i|t_j}, \sigma_{i|t_j}|X_i)
\tag{3.1}
\end{equation*}
Moving forward, we express the posteriors as the product of the likelihood with the prior while omitting the normalizing constant:
\begin{equation*}
\centering
 h(t_j) \propto \prod_{i=1}^n P(X_i | \mu_{i|t_j},\sigma_{i|t_j}) P(\mu_{i|t_{j-1}},\sigma_{i|t_{j-1}} | X_i)
\tag{3.2}
\end{equation*}
where $P(X_i|\mu_{i|t_j},\sigma_{i|t_j})$ is the p.d.f. of the Gaussian (Normal) distribution with parameters derived at each time step using the MLE method (see derivation details in \textbf{Appendix A}), with the likelihood function $L$ at $t_j$ given by: \vspace{-.4cm}
\begin{equation*}
L(X; \hat{\mu},\hat{\sigma} | t_{0:j}, D = 1) = \prod_{i=1}^{N_{t_{0:j}}} \frac{1}{\sqrt{2\pi\sigma^2}} \mathrm{e}^{-\frac{1}{2} \left(\frac{x_i-\mu}{\sigma} \right)^2}
\end{equation*}
in which we utilize the whole history $ \{x_i\}_{i=1}^{N_{t_{j}}}$ of past event occurrences up to time step $t_j$ for estimating the parameters. Therefore, the likelihood component for a single covariate $X$ can be estimated by:
\begin{equation*}
\centering
P(X| \mu_{t_j}, \sigma_{t_j}) =  \prod_{i=1}^{N_{t_{0:j}}} \frac{1}{\sqrt{2\pi\ \hat{\sigma}^2_{t_{0:j}}}}\mathrm{e}^{-\frac{1}{2} \left(\frac{x_i-\hat{\mu}_{t_{0:j}}}{\hat{\sigma}_{t_{0:j}}} \right)^2}
\tag{3.3}
\end{equation*}
and for avoiding arithmetic underflow we can sum the log-likelihood of each covariate, such that: 
\begin{align}
P(X|\mu_{t_j},\sigma_{t_j}) \, &\leftarrow 1 + \displaystyle\sum_{i=1}^{N_{t_{0:j}}} \log\left(\frac{1}{\sqrt{2\pi\hat{\sigma}^2_{t_{0:j}}}}\right) \nonumber \\
&\quad \,\,\,\,\,\,\,\,\,\,\,\,\, - \frac{1}{2}\left(\frac{x_i-\hat{\mu}_{t_{0:j}}}{\hat{\sigma}_{t_{0:j}}}\right)^2
\tag{3.4}
\end{align}

By incrementing the estimated probability by 1, we mitigate the need for multiplication by a fraction when employing the current posterior probability as the prior component in the subsequent time step.

To simplify the notation and enhance the reader's understanding of our methodology, we present the hazard function for a single covariate modeled at each time step using the recursive function:
\vspace{.075cm}
\begin{align*}
h(t_{j}) = P(X| \hat{\mu}_{t{_{j}}}
,\hat{\sigma}_{t{_{j}}}) \, P(\hat{\mu}_{t{_{j-1}}},\hat{\sigma}_{t_{j-1}}|X) \nonumber \\
\end{align*}
\vspace{-.95cm}
\begin{equation*}
=> h(t_{j}) := P(X| \hat{\mu}_{t{_{j}}},\hat{\sigma}_{t{_{j}}}) \, h(t_{j-1})
\end{equation*}

\vspace{.075cm}
where $j>0$, with $h(t_0)=1$ being a non-informative prior, and
for $n$ covariates: \vspace{-.1cm}
\begin{equation*}
\centering
h(t_j) = \prod_{i=1}^n P(X_i | \mu_{i|t_j},\sigma_{i|t_j}) \, h_i(t_{j-1})
\tag{3.5}
\end{equation*}
and once again for avoiding arithmetic underflow, we sum
the components of each covariate: 
\begin{equation*}
h(t_j) \, \leftarrow \sum_{i=1}^n log \left(P(X_i|\mu_{i|t_j},\sigma_{i|t_j})  \, h_i(t_{j-1}) \right)
\tag{3.6}
\end{equation*}
Following the estimation of hazard rates at each time step, we calculate the aggregate survival function by:
\begin{align*}
\hat{S}_{base}{(t_j)}=1-\left(\frac{h\left(t_j\right)}{\max{\left(h\right)}}\right)
\tag{3.7}
\end{align*} \vspace{-.28cm}
where $h=\left\{h\left(t_j\right)\right\}_{j=0}^J$
\vspace{.2cm}

\vspace{-.22cm}
\begin{figure}[htb]
\centering
\includegraphics[width=3.35in,height=3.1in]{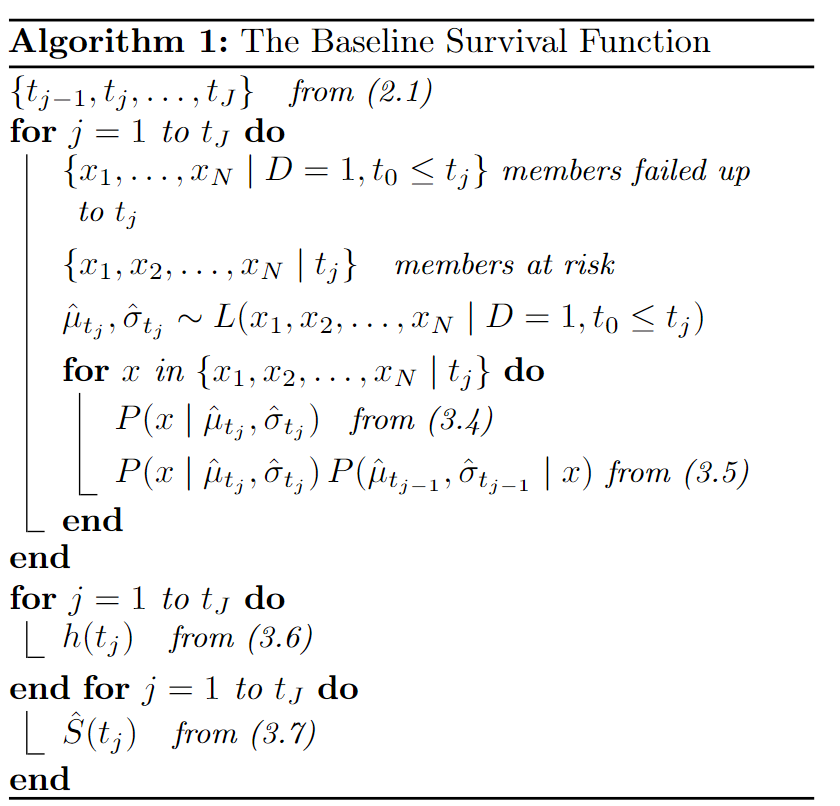}
\vspace{-.1cm}
\end{figure}
\vspace{-.3cm}

\section{Survival Scores and the Individual Survival Curves}
\vspace{-.22cm}
In this section, we introduce the notions of individual survival scores and survival curves. While the baseline hazard governs the time-to-event distribution of a given population, survival scores capture variations in survival based on individual member characteristics. Specifically, we generate the survival scores for new members and combine them with the baseline hazard for estimating their survival probabilities for the duration of a study. \vspace{-0.25cm}
\subsection{Survival Scores} \vspace{-0.25cm}
For generating the survival scores, we fit a binary classifier on the training data ${\{(x_i,D_i)}\}_{i=1}^{N_{train}}$ without taking the time covariate into consideration. Particularly we are interested in the weights vector, and how each weight relates with its respective covariate. 

\textbf{Assumption 3:} \textit{Proportional Hazard Rates - If the values of the covariates $X$ remain fixed, and subject A has a greater predicted probability of survival than subject B at $t_1$, then this relationship must hold true for all other time points up to the conclusion time $J$}. \vspace{.15cm}
\begin{align*}
&\small P\left(\widehat{S_A} | t_1\right) > P\left(\widehat{S_B} | t_1\right) \,\, \land \,\, \small P\left(\widehat{S_A} | t_2\right) > P\left(\widehat{S_B} | t_2\right) \\ 
& \land \ldots \,\, \ldots \land \small \, P\left(\widehat{S_A} | t_J\right) > P\left(\widehat{S_B} | t_J\right)
\end{align*}
\normalsize
Hence the hazard (and survival) ratios at each time step must be equal:

$h_{AB|t_1}=h_{AB|t_2} =  ... \, = h_{AB|t_J}$ \, \small where \, $h_{AB|t_j}= \frac{h_{A|t_j}}{h_{B|t_j}}$
\normalsize \\ 
\vspace{-.05cm} 
\\
\textbf{Proposition 1:}\textit{ Training a binary classifier, the resulting scalar product of the classifier's pretrained weights with a new test sample yields distinct survival scores for each event.}


\textbf{Proof 1:} \textit{If two distinct covariates share the same weight, if 
$x_1 \neq x_2 => S(x_1|D=0) \neq S(x_2|D=1)$}

Using the sigmoid activation function for each binary outcome:
\vspace{-.2cm}
\begin{equation}
S\left(x_1\middle| D=0\right)= \frac{1}{1+e^{-\left(w_1x_1\right)}} =  y \hspace{1em}
\end{equation}
\vspace{-.2cm}
and
\vspace{-.2cm}
\begin{equation}
S\left(x_2\middle| D=1\right)= \frac{1}{1+e^{-\left(w_2x_2\right)}} = 1-y \,\,\,\,\,\,
\end{equation} where  $y \, \in \{0,1\}.$

Replacing equation (1) into equation (2) and letting $w_1 = w_2$:
\\ 
\\
 $S\left(x_2\middle| D=1\right)=$\  \large$\frac{1}{1+e^{-\left(wx_2\right)}}$\normalsize \  $=\ 1-$ \large $\frac{1}{1+e^{-\left(wx_1\right)}}$
 \normalsize

 $=>$ \large \ \  $\frac{1}{1+e^{-\left(wx_2\right)}}$ \normalsize $+$ \large \ $\frac{1}{1+e^{-\left(wx_1\right)}}$ \normalsize $=1$ 
 \vspace{-.2cm}
 \\
 
Multiplying by $(1 + e^{-wx_2}) * (1 + e^{-wx_1})$:
\vspace{.25cm}
\\
$=>$ $\ \frac{\left(1+e^{-\left(wx_2\right)}\right)\left(1+e^{-\left(wx_1\right)}\right)}{1+e^{-\left(wx_2\right)}}$  $+$\ $\frac{\left(1+e^{-\left(wx_2\right)}\right)\left(1+e^{-\left(wx_1\right)}\right)}{1+e^{-\left(wx_1\right)}}\ 
\small
\\ 
=\left(1+e^{-\left(wx_2\right)}\right)\left(1+e^{-\left(wx_1\right)}\right)$
\normalsize

$=> 2 + e^{-wx_2} + e^{-wx_1} = \left(1+e^{-\left(wx_2\right)}\right)\left(1+e^{-\left(wx_1\right)}\right)$ \vspace{.125cm}
\\ 
expanding the RHS:

$=> 2 + e^{-wx_2} + e^{-wx_1} = 1 + e^{-wx_2} + e^{-wx_1} + e^{-w(x_1+x_2)}$ \,\, 

and removing the common terms: \par

\large
$e^{-w\left(x_1+x_2\right)}$ \normalsize $=1  => -w\left(x_1+x_2\right)=0
\vspace{0.1cm}
\\ \
=> x_1=-x_2\ \bm{{\Longleftrightarrow\ S(x_1|D=0) \neq S(x_2|D=1)}}$

which concludes Proof 1. Thus, if two (or more) covariates share the same weights, their individual covariate observations will consistently yield significantly different results if the event of interest differs. Based on the above proof, we define the individual survival score to be the scalar product of the test set with the pretrained weights from the training set, i.e.: $r = w^Tx_t$ where $x_t$ denotes a new (out-of-population) member. 
\vspace{-0.25cm}
\subsection{Survival Curves} \vspace{-0.25cm}
For generating the survival curves, we perform a data transformation that results in both the baseline hazard and the survival scores to be of the same magnitude. For the vector of survival scores \underline{$r$}:
\begin{align*}
\ r_{i}=\  \frac{|\min{\left(\underline{r}\right)|}+\ r_{i}}{\left|\min{\left(\underline{r}\right)}\right|+\ \max{\left(\underline{r}\right)}}
\tag{4.2.1}
\end{align*}
With the individual survival probabilities being the exponential product of each individual survival score with the baseline survival probabilities. 
\begin{align*}
\centering
\widehat{\underline{\ S_\ }}=\exp{\left(\left(1+\underline{r}\right)\ast\underline{{\hat{S}}_{base}}\right)}
\tag{4.2.2}
\end{align*}
And dividing by the max value for generating a range between [0,1], yields the individual survival curves. 
\begin{align*}
    \widehat{S_{i}}=\ \widehat{S_{i}}\ / max(\underline{\hat{S}})
    \tag{4.2.3}
\end{align*}
As it can be seen from Figure 2, having a survival (risk) score that does not vary with time results in survival curves that do not cross each other, as well as constant hazard rates. Therefore, \textbf{Assumption 3} is satisfied. Compared with the KM survival probabilities and the proposed baseline hazard, the individual survival curves offer a versatile solution since they can separate different members, and allow to cross (or not) the baseline hazard depending on the survival score of each member. 
\vspace{-.05cm}
\subsection{Alternating Baselines}
\vspace{-.21cm}
The framework we propose is versatile and can be used with various baseline hazard/survival functions. For instance, it can be adjusted to feature the KM estimator for the baseline survival function, i.e.: 

\vspace{-0.25cm}
\begin{align*}
\centering
\hat{S}_{base}(t_j) = \prod_{t_j \leq t_J}^{} 1\ -\left(\frac{d_{t_j}}{n_{t_j}}\right)
\end{align*}
\vspace{-.3cm}
\begin{align*}
\widehat{\underline{\ S_\ }}=\exp{\left(\left(1+\underline{r}\right)\ast\underline{{\hat{S}}_{base}}\right)}
\end{align*}
And dividing by the max value for generating again a range between [0,1], once again yields the individual survival curves. 
\begin{align*}
    \widehat{S_{i}}=\ \widehat{S_{i}}\ / max(\underline{\hat{S}})
\end{align*}
Therefore, we still utilize the individual survival scores to estimate survival probabilities for each member. However, we use a different baseline for modeling the time-to-event distribution. As demonstrated in the next section, since the survival scores do not change, the discrimination performance of the model remains constant. However, altering the baseline survival can improve a model's calibration performance, especially when a specific baseline is better suited for a particular application, as demonstrated in the next section.

\section{Experiments and Results}
\vspace{-0.15cm}

\begin{figure*}[h]
\centering
  \includegraphics[width= 6.765in,height=2.71in]{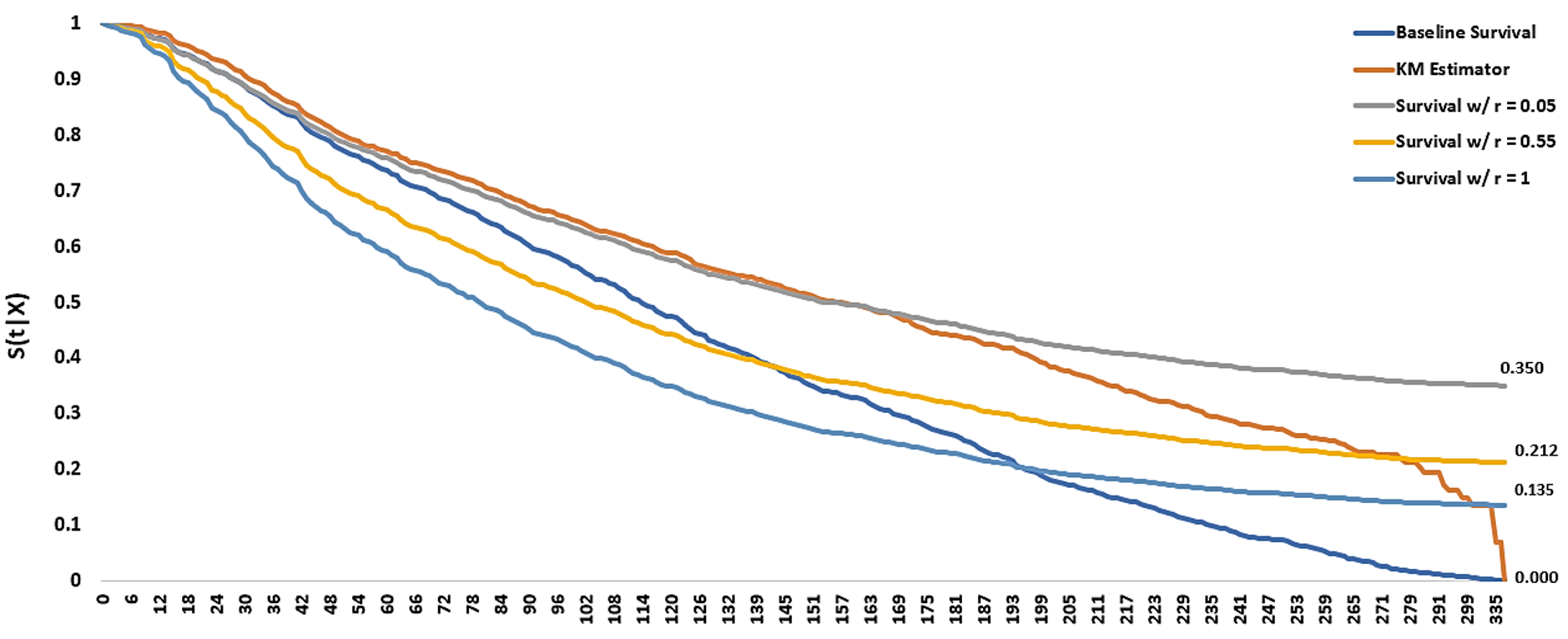}
  \vspace{-0.4cm}
  \caption{\small \textbf{Application of our proposal and the KM’s estimator in the METABRIC dataset. The baseline survival curve of our model starts at 1 and monotonically decreases to 0. The risk scores of each participant determine the shape of their respective survival curves. In this paradigm, the distributions for participants with varying risk scores are presented for illustrative purposes. Members with lower risk scores exhibit greater likelihood of survival than members with high-risk scores. As the proportionality assumption holds, and the risk scores do not vary with time, the participants’ survival curves do not cross each other, although
they can cross the baseline survival.} \normalsize}

\end{figure*}

\subsection{Evaluation Metrics}
\vspace{-0.2cm}
For evaluating the success of our proposal, we employ the following metrics, and refer the readers to \citet{kvamme2019time}, \citet{steyerberg2010assessing}, \citet{ishwaran2008random} for a more detailed description.

\textbf{Time-dependent Brier Score:} A measure for both calibration and discrimination. It  evaluates the predictive accuracy of the model by comparing the prediction (event occurring or staying event-free) over a fixed period of time, with the censoring weight $\hat{G}\left(t\right)$ estimated by the Kaplan Meier Estimator. 


\begin{align*}
BS(t) &= \frac{1}{N} \sum_{i=1}^N \left(\frac{\hat{S}(t|x_i)^2 I\{T^* \leq t, D_i = 1\}}{\hat{G}(T^*_i)} \right. \\
&\quad \left. \,\,\,\,\,\,\,\, + \,\, \frac{(1 - \hat{S}(t|x_i))^2 I\{T^*_i>t\}}{\hat{G}(t)} \right)
\tag{5.1.1}
\end{align*}

\vspace{-.2cm}
Equation (5.1.1) can be integrated between a time interval, yielding the Brier score over that interval.
\begin{align*}
IBS=\frac{1}{t_2-t_1}\int_{t_1}^{t_2}{BS\left(s\right)ds}
\tag{5.1.2}
\end{align*}

\vspace{-.2cm}
\textbf{Concordance index:} C-index is the most frequently used metric in Survival Analysis and parallels the accuracy of a classifier model. It ranks the ordering of times amongst all possible pairs of the participants in a study. On test data C-index is estimated as the fraction of pairs whose predicted death times have correct order compared to their true death times. In this work we utilize the version by \cite{antolini2005time} since it can be applied equally well in both proportional and non-proportional survival models.

\begin{align*}
C=\frac{\sum_{i\neq j}{I\left\{n_i<n_j\right\}I\left\{T^*_i>T^*_j\right\}D_j\ }}{\sum_{i\neq j}{I\left\{T^*_i>T^*_j\right\}D_j}}\ 
\tag{5.1.3}
\end{align*}

\textbf{Time-Dependent AUC:} The ROC curve can be applied to survival data by estimating the sensitivity and positivity as time-dependent quantities. Hence, it can be used to quantify how well a model can distinguish the members who fail by a given time \small $\left(T^*_i\le t\right)$ \normalsize from members that fail after \small $(T^*_i>t)$\normalsize. In the equation below, \small $\hat{f}\left(x_i\right)$ \normalsize is defined as a risk score at time $t$, or \small $1-P\left(\hat{S}\ | t\right)$, and $\omega_i$\normalsize \, is the inverse probability of censoring weights computed using the KM estimator. \\ \\ 
$\hat{AUC}(t) =$ \vspace{-.2cm}
\begin{equation*}
\frac{\sum_{i=1}^N \sum_{j=1}^N \, I(T^*_j > t) \, I(T^*_i \leq t) \, \omega_i \, I(\hat{f}(x_j) \leq \hat{f}(x_i)) }{\sum_{i=1}^N I(T^*_i > t) \,\, \sum_{i=1}^N I(T^*_i \leq t)\omega_i}
\tag{5.1.4}
\end{equation*}

\textbf{Mean of Time-Dependent AUC:} Yields a summary measure of the time-dependent AUC over the time range \small $(\tau_1,\tau_2)$\normalsize. \par

{\small
\begin{align*}
\overline{AUC}\left(\tau_1,\tau_2\right)=\frac{1}{\left(\hat{S}\left(\tau_1\right)-\hat{S}\left(\tau_2\right)\right)^2}\ \ast\int_{\tau_1}^{\tau_2}{\widehat{AUC}\left(t\right)d\hat{S}\left(t\right)}
\tag{\normalsize{5.1.5}}
\normalsize
\end{align*}}%

\subsection{Application and Results}
\vspace{-.35cm}
\begin{table}[h]

\centering
\caption{\textbf{Dataset Description}} \label{sample-table}
\vspace{0.15cm}
\begin{tabular}{lccc}
\hline
\textbf{\small } &\textbf{\small Size} &\textbf{\small Covariates}  &\textbf{\small Censored \%} 
\\
\hline \\ 
SUPPORT             &8,873 &14 &32 \%\\ 
METABRIC            &1,904 &9  &42 \%\\
GBSG                &2,232 &7  &43 \%\\
FLCHAIN             &6,524 &8  &70 \%\\
\hline
\end{tabular}
\end{table}

\vspace{-.15cm}

\begin{table*}[h!]
\centering
\caption{\textbf{Antolini's C-index $(\uparrow)$}} 
\begin{tabular}{ccccc}
\\
\hline
\textbf{\small Method } &\textbf{\scriptsize SUPPORT} &\textbf{\scriptsize FLCHAIN} &\textbf{\scriptsize METABRIC} &\textbf{\scriptsize GBSG} 
\\
\hline \\

BS              &$0.560 \pm 0.0074$  &$0.788 \pm 0.0075$ &$0.630 \pm 0.019$ &$0.665 \pm 0.013$ \\
BS w/KM         &$0.560 \pm 0.0074$  &$0.788 \pm 0.0075$ &$0.630 \pm 0.019$ &$0.665 \pm 0.013$ \\
Cox Net         &$0.572 \pm 0.0089$  &$0.792 \pm 0.0053$ &$0.636 \pm 0.018$ &$0.663 \pm 0.011$\\
RSF             &$0.633 \pm 0.0090$  &$0.787 \pm 0.0058$ &$0.632 \pm 0.020$ &$0.649 \pm 0.013$\\
EST             &$0.597 \pm 0.0081$  &$0.779 \pm 0.0094$ &$0.639 \pm 0.014$ &$0.656 \pm 0.021$\\
Cox Time        &$0.626 \pm 0.076$  &$0.766 \pm 0.0170$ &$0.661 \pm 0.016$ &$0.669 \pm 0.020$ \\
SVM             &$0.575 \pm 0.0082$  & $0.792 \pm 0.0043$ &$0.639 \pm 0.014$ &$0.667 \pm 0.011$ \\
\hline
\end{tabular}
\end{table*}
\vspace{-.1cm}

\begin{table*}[h!]
\centering
\caption{\textbf{Avg of Time-Dependent AUC $(\uparrow)$}} \label{sample-table}
\begin{tabular}{ccccc}
\\
\hline
\textbf{\small Method} &\textbf{\scriptsize SUPPORT} &\textbf{\scriptsize FLCHAIN} &\textbf{\scriptsize METABRIC} &\textbf{\scriptsize GBSG} 
\\
\hline \\
BS              &$0.527 \pm 0.013$  &$0.767 \pm 0.082$ &$0.631 \pm 0.036$ &$0.720 \pm 0.015$ \\
BS w/KM         &$0.527 \pm 0.013$  &$0.767 \pm 0.081$ &$0.631 \pm 0.036$ &$0.720 \pm 0.015$ \\
Cox Net         &$0.561 \pm 0.016$  &$0.770 \pm 0.076$ &$0.645 \pm 0.043$ &$0.716 \pm 0.011$\\
RSF             &$0.572 \pm 0.006$ &$0.777 \pm 0.069$ &$0.601 \pm 0.037$ &$0.685 \pm 0.026$ \\
EST             &$0.571 \pm 0.017$  &$0.760 \pm 0.077$ &$0.668 \pm 0.040$ &$0.705 \pm 0.030$ \\
Cox Time        &$0.571 \pm 0.020$  & $0.756 \pm 0.069$ &$0.558 \pm 0.045$ &$0.731 \pm 0.024$ \\
SVM             &$0.601 \pm 0.017$  & $0.777 \pm 0.072$ &$0.687 \pm 0.029$ &$0.721 \pm 0.011$ \\
\hline
\end{tabular}
\end{table*}
\vspace{.2cm}

\begin{table*}[h!]
\centering
\caption{\textbf{Integrated Brier Score $(\downarrow)$}} \label{sample-table}
\begin{tabular}{ccccc}
\\
\hline
\textbf{\small Method} &\textbf{\scriptsize SUPPORT} &\textbf{\scriptsize FLCHAIN} &\textbf{\scriptsize METABRIC} &\textbf{\scriptsize GBSG} 
\\
\hline \\
BS              &$0.211 \pm 0.002$  &$0.281 \pm 0.005$ &$0.179 \pm 0.004$ & $0.191 \pm 0.002$ \\
BS w/KM         &$0.210 \pm 0.003$  &$0.147 \pm 0.004$ &$0.177 \pm 0.004$ & $0.192 \pm 0.002$ \\
Cox Net         &$0.205 \pm 0.003$  &$0.121 \pm 0.004$ & $0.160 \pm 0.005$ &$0.181 \pm 0.003$ \\
RSF             &$0.191 \pm 0.003$  &$0.119 \pm 0.004$ &$0.170 \pm 0.006$ &$0.184 \pm 0.007$ \\
EST             &$0.203 \pm 0.002$  &$0.134 \pm 0.002$ &$0.165 \pm 0.004$ &$0.186 \pm 0.004$ \\
Cox Time        &$0.192 \pm 0.002$ &$0.130 \pm 0.016$ &$0.155 \pm 0.006$ &$0.177 \pm 0.005$ \\
SVM             &$0.264 \pm 0.009$  & $0.136 \pm 0.004$ &$0.241 \pm 0.004$ &$0.226 \pm 0.006$ \\
\hline
\end{tabular}
\end{table*}
\vspace{-.2cm}
We apply our proposal to four real-world data sets namely, SUPPORT, METABRIC, FLCHAIN, and GBSG, as presented in \citet{kvamme2019time}. We compare the results of our model (BS) with Random Survival Forest - \citet{ishwaran2008random}, classical Cox Model - \citet{cox1972regression} with regularization, Survival SVM \citet{polsterl2016efficient}, Gradient Boosting Cox (EST), and because of computational constraints, Cox-Time \citet{kvamme2019time} since it achieves state-of-the-art results among the DL survival models.We split the data sets into 80\% training data and 20\% test data (performing 5-fold cross-validation), with the same random seeds. The continuous covariates were standardized, and the categorical covariates were one-hot-encoded. For the conventional models, their parameters were left at default, and for Cox-Time the learning rate was determined using the finder from \textbf{\href{https://github.com/havakv/pycox}{pycox}}, with the remaining hyperparameters as recommended. Cox Time was trained for 200 epochs.

For evaluating the performance of each model, we utilize the metrics presented as presented in the beginning of the section. For the AUC (eq. 5.1.4) of the ROC curve, since it requires a single risk estimate $\hat{f}(x_i)$, we use the probability of failure at the last time step for each participant. This ensures a fair comparison with models that do not generate time-dependent risk scores. The mean AUC is evaluated for weekly (7-day) periods for 90 days after each study has commenced.

Overall, our approach, of decomposing the survival function into a baseline survival and individual risk scores, consistently achieves competitive results across all three evaluation metrics. Demonstrating robust performance, our proposal exhibits stable results when handling datasets with diverse characteristics. Nevertheless, in terms of calibration (IBS), when our model records less favourable results, we can experiment with different survival functions. For instance, incorporating the Kaplan-Meier estimator as the baseline survival notably improves calibration in the case of FLCHAIN. It's essential to highlight that changing the baseline survival function does not impact the discrimination performance, as it is governed by the individual survival scores.

While alternative models may, at times, achieve more competitive results, their performance lacks the robustness demonstrated by our proposal across all metrics and datasets. For example, SVM exhibits strong discrimination performance but demonstrates inferior calibration results. Other models produce more volatile results, with the exception of Cox Time, which exhibits better and more stable results overall; although it requires significantly more time for training and calibrating its hyperparameters. 

In summary, our proposed scheme stands out as a robust choice, providing a stable foundation for Survival Analysis across diverse datasets. Furthermore, users are able to experiment with different baseline survival functions and models to potentially achieve superior performance. For instance, a possible combination, will be to employ the KM estimator as the baseline survival function and a DNN for learning underlying patterns in the form of survival scores.

\vspace{-.07cm}
\section{Conclusions}
\vspace{-.09cm}
In this work, we introduced a novel framework for inferring the individual survival functions of a population. Our framework comprises two key components: a Bayesian approach to estimate the baseline hazard and a methodology for deriving the survival scores/curves of each member. The baseline hazard effectively manages censoring, while the survival scores offer flexibility by incorporating covariates to generate survival curves for each member.
 
Despite making certain assumptions, our scheme demonstrates robustness as it still performs well even when some of these assumptions are not strictly met. This is attributed to the independence between the baseline hazard and the covariate model, distinguishing it from other proportional hazards models. Additionally, our proposal offers computational efficiency and requires minimal fine-tuning, while delivering results comparable to other state-of-the-art SA models.
Finally, the proposed framework is versatile and can be adapted to alternative baseline survival functions that may better suit specific scenarios, or different models that generate heterogeneous risk scores. 

The work performed in this paper sets the foundation for future work that we intend to carry out. In particular, we aim to model the survival scores of each member using the time covariate, further enhancing the framework's flexibility and predictive capabilities, by generating dynamical survival curves. 

\bibliography{references}

\newpage

\onecolumn
\section*{Appendix A} 
\label{sec:Appendix}
\vspace{-.2cm}
\subsection*{A.1 Deriviation and Optimization of the Baseline Survival Function}
\vspace{-.2cm}
\subsubsection*{A.1.1 Motivation}
\vspace{-.2cm}
We employ a Bayesian approach to hazard estimation for each covariate. Our method involves a recursive function, which is the product of the likelihood (reflecting the probability of observing the current set of observations) and the posterior from the previous step, serving as the prior. Probabilities are estimated using the pdf of the Gaussian Distribution, with parameters determined at each time step through the Maximum Likelihood Estimation (MLE) method. This choice is motivated by our interest in parameters maximizing the likelihood given the occurrence of the event. We aim to estimate the probability of observing a new data point with these parameters. Observations closer to the sample mean, calculated from past data for each covariate, yield a higher probability of occurrence for the specific observation. Consequently, a higher probability of the event at the current time step increases the hazard rate and decreases the survival function (1 – hazard rate). 
\vspace{-.2cm}
\subsubsection*{A.2.1 Derivation of the sample parameters}
\vspace{-.2cm}
For a single covariate that follows a Gaussian distribution $X\sim N(\mu,\sigma)$ the likelihood of observing the particular sample can be expressed as:
\begin{equation*}
\centering
L(\mu,\sigma | \{x_i\}_{i=1}^N) = \prod_{i=1}^N \frac{1}{\sqrt{2\pi\sigma^2}} \, exp \left(-\frac{1}{2} \left(\frac{x_i - \mu}{\sigma}\right)^2\right) 
\end{equation*}

In our work we are interested in estimating the parameters that maximize the likelihood of observing the particular sample at each step, from the beginning of the study up to and including $t_j$. Hence the likelihood can be expressed as:
\begin{equation*}
\centering
L(\mu,\sigma| \{ x_i |\tiny{t_{0:j};D=1}\}_{i=1}^N) =  \prod_{i=1}^{N_{t_{0:j}}} \frac{1}{\sqrt{2\pi\sigma^2}} \, exp \left(-\frac{1}{2} \left(\frac{x_i - \mu}{\sigma}\right)^2\right) 
\end{equation*}
From now on we denote the likelihood function asc $L$ for avoiding notation burden:

\begin{equation*}
\centering L = \frac{1}{(\sqrt{2\pi \sigma^2})^{N_{t_{0:j}}}} \, exp \left(-\frac{1}{\sigma^2} \sum_{i=1}^{N_{t_{0:j}}} (x_i - \mu)^2 \right)
\end{equation*}

gathering all the constant terms in $C$, and taking the log-likelihood:
\begin{equation*}
log(L) = - N_{t_{0:j}} log(\sigma) - \frac{1}{2\sigma^2} \sum_{i=1}^{N_{t_{0:j}}} (x_i - \mu)^2 + C
\end{equation*}

and taking the partial derivatives with respect to µ:
\begin{equation*}
\centering
\frac{\partial log(L)}{\partial \mu} = - \frac{1}{2\sigma^2} \sum_{i=1}^{N_{t_{0:j}}} (2)(-1)(x_i - \mu)^2 = \frac{1}{2\sigma^2} \sum_{i=1}^{N_{t_{0:j}}} x_i - ({N_{t_{0:j}}} * \mu) = 0
\end{equation*}

setting equal to 0 and solving for µ:
\begin{equation*}
\hat{\mu}_{t_j} = \frac{1}{N_{t_{0:j}}} \sum_{i=1}^{N_{t_{0:j}}} x_i
\end{equation*}
\\
the partial derivative with respect to $\sigma$:
\begin{equation*}
\centering
\frac{\partial log(L)}{\partial \sigma} = -\frac{N_{t_{0:j}}}{\sigma} - \frac{1}{2} \left(-\frac{2}{\sigma^3}\right) \sum_{i=1}^{N_{t_{0:j}}} (x_i - \mu)^2 = 0 
\end{equation*}

\begin{equation*}
=> \frac{1}{\sigma^3} \sum_{i=1}^{N_{t_{0:j}}} (x_i - \mu)^2 = \frac{N_{t_{0:j}}}{\sigma} => \hat{\sigma}_{t_j} = \sqrt{\frac{\sum_{i=1}^{N_{t_{0:j}}} (x_i - \mu)^2}{N_{t_{0:j}}}}
\end{equation*}

\end{document}